\documentclass[runningheads]{llncs}

\usepackage{eccv}

\usepackage{eccvabbrv}

\usepackage{graphicx}
\usepackage{booktabs}

\usepackage[mathscr]{euscript}

\def\bF{{\bf F}}
\def\bC{{\bf C}}
\def\bI{{\bf I}}
\def\bG{{\bf G}}
\def\bV{{\bf V}}
\def\bM{{\bf M}}
\def\bD{{\bf D}}
\def\bE{{\bf E}}
\def\bK{{\bf K}}
\def\matR{{\mathbb{R}}}

\usepackage{amsmath}
\usepackage{amssymb}
\usepackage{mathtools}
\usepackage{multirow}
\usepackage{siunitx}
\usepackage{xspace}
\usepackage{bbding}
\usepackage{makecell}

\usepackage[accsupp]{axessibility}  %

\usepackage[pagebackref,breaklinks,colorlinks,citecolor=eccvblue]{hyperref}

\usepackage{orcidlink}

\def\OURS{{RGM}\xspace}

\begin{document}

\title{RGM: A Robust Generalizable Matching Model\thanks{This work was supported by National Key R\&D Program of China (No.\ 20\-22\-ZD\-01\-18\-7\-00). 
SZ's contribution was made when he was with Zhejiang University.
BL’s participation was in part supported by National Natural Science Foundation of China (No.\ 62\-00\-13\-94). SZ, XS
contributed equally. HC, BL are the corresponding authors.}
} 

\titlerunning{Robust Generalizable Matching}

\author{
Songyan Zhang$^{2}$,~~
Xinyu Sun$^1$, ~~
Hao Chen$^1$, ~~
Bo Li$^3$, ~~ Chunhua Shen$^1$
}

\authorrunning{Zhang et al.}

\institute{
$^1$Zhejiang University, China ~~
$^2$Nanyang Technological University, Singapore ~~
$^3$Northwestern Polytechnical University, China 
}

\maketitle
\begin{figure}[!htb]

\begin{minipage}[h]{0.67\linewidth}
    \centering
    \includegraphics[width=\linewidth]{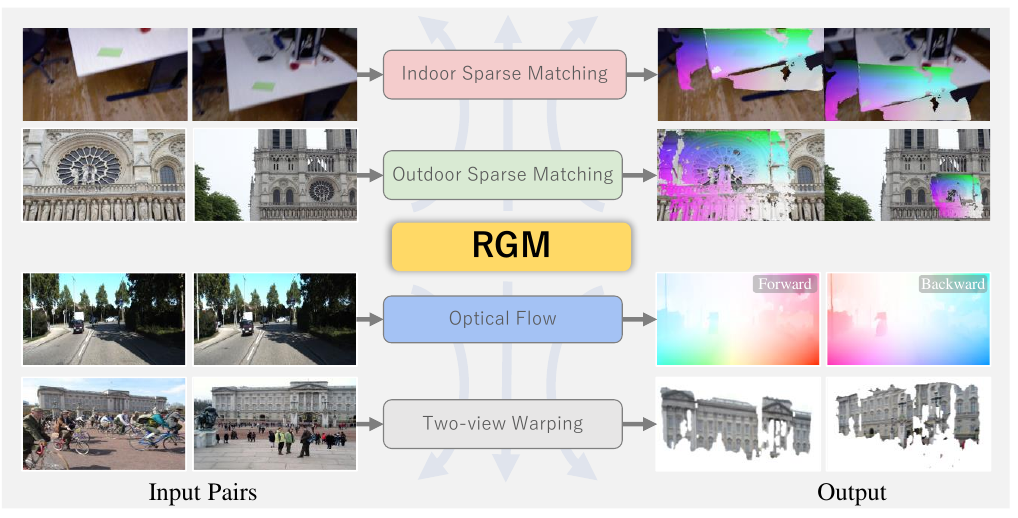}
    \tiny{(a) Generalizable matching of our RGM.}
\end{minipage}
\hfill
\begin{minipage}[h]{0.31\linewidth}
    \centering
    \includegraphics[width=\linewidth]{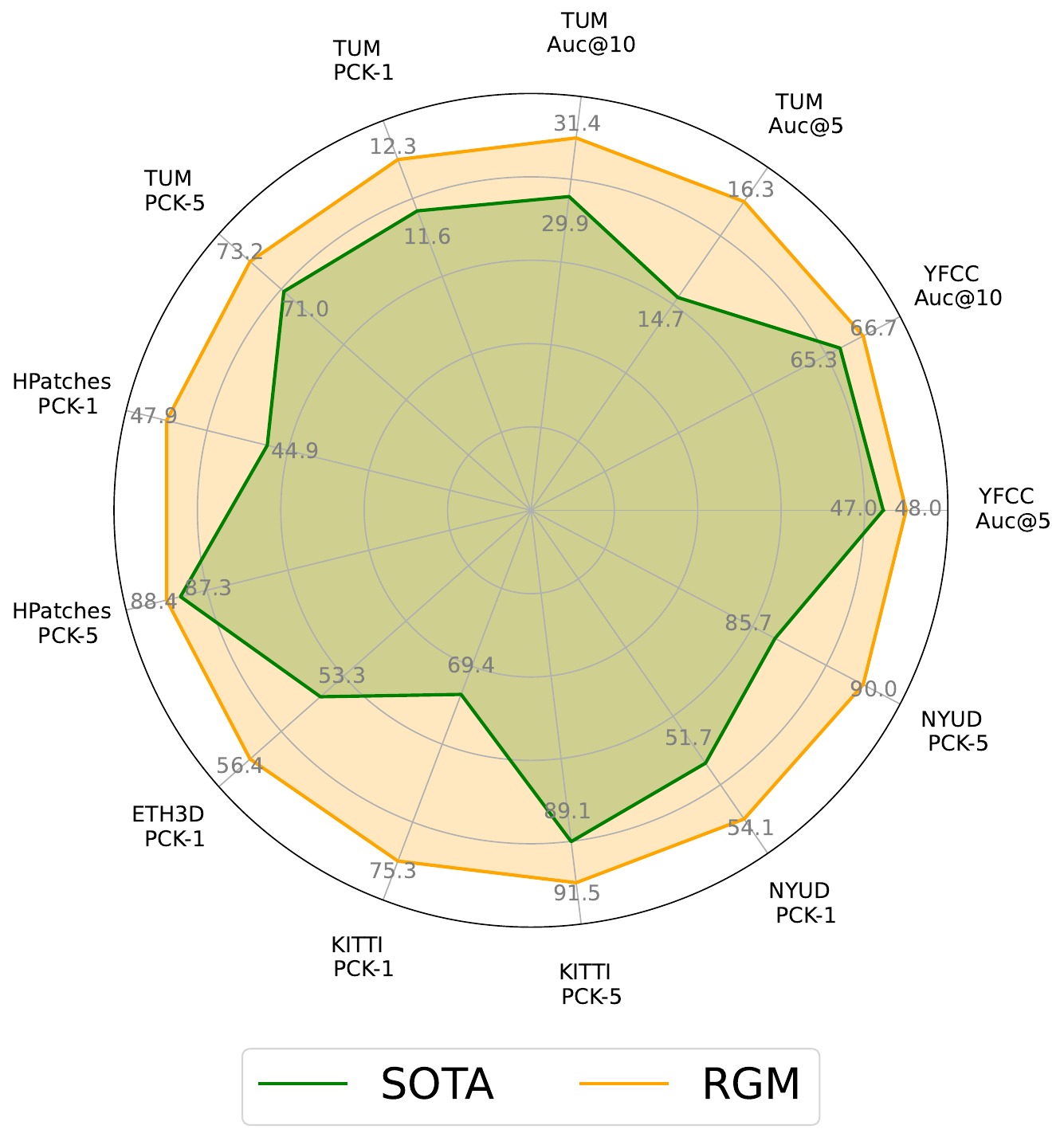}
    \tiny{(b) Generalization comparisons.}
\end{minipage}
\caption{
    \textbf{RGM overview.} We propose a \textbf{r}obust \textbf{g}eneralizable \textbf{m}atching model, termed RGM. (a): RGM shows generalizable performance for both sparse and dense matching covering indoor and outdoor scenarios. (b): Our RGM achieves excellent generalization performance, outperforming previous SOTA methods by %
    a clear 
    margin.
}
\label{fig:Teaser}
\end{figure}

\begin{abstract}
  Significant progress has been made for sparse and dense matching with specialized architectures and task-specific datasets that vary substantially, but the zero-shot generalization capacity is less satisfactory. Recently, learning all-paired correspondence jointly with an uncertainty-based sparsification module makes a significant step for unifying sparse and dense matching, leading to an appealing possibility of improved generalizable capacity by taking advantage of the abundant diversity in task-specific datasets with a unified matching model. 
  
  However, the significant domain gap within data from different sources as well as the potential conflict in dense matching and its sparsification make it a challenge to realize this expected capacity. To tackle this challenge, we propose a robust generalizable matching model, termed RGM. A decoupled learning strategy is explored to learn the matching and uncertainty-based sparsification in a two-stage hierarchical manner, learning the uncertainty from the well-learned matches. This enables us to alleviate the mutual interference for better learning efficiency and explore the scaling law of different datasets for different tasks separately. To enrich the data diversity, we collect datasets from tasks of dense optical flow, sparse local feature matching, and a synthesized dataset with a similar displacement distribution to real-world data, reaching 4 million data pairs in all.
  
  Extensive experiments demonstrate our state-of-the-art generalization capacity for zero-shot matching and pose estimation, outperforming previous methods by a large margin.

\end{abstract}

\section{Introduction}

Correspondence matching is a fundamental task in computer vision with various applications %
such as 
Simultaneous Localization and Mapping (SLAM), multi-view geometry estimation, and image editing. Due to the specific requirements of different applications, 
recent learning-based matching works are mainly categorized into two branches: \textit{sparse} and \textit{dense} matching. Benefiting from the accumulating public datasets~\cite{megadepth, scannet, things3d, sintel}, learning-based approaches have witnessed remarkable improvement for particular matching tasks such as dense matching for optical flow estimation \cite{raft, flowformer, pwcnet, gmflow}, stereo matching \cite{EDNet, AANet, psmnet, ACVNet}, and sparse matching for geometry estimation \cite{loftr, aspanformer, pats, edstedt2023dkm}. However, the poor generalization capacity of matching models still demands further exploration.

The difference in sparse and dense matching is primarily twofold which causes poor generalization performance. Firstly, in the context of dense matching like optical flow estimation, the image pairs to be matched typically have limited changes of viewpoint with a relatively small temporal interval while the image pairs for sparse feature matching normally exhibit more significant changes in viewpoint with various image properties. Besides, task-specific datasets exhibit significant domain gaps in both real-world and synthesized data across indoor and outdoor scenarios. Secondly, dense matching, 
\textit{e.g.}, optical flow estimation, is required to provide all-paired matching even in occluded regions, while sparse feature matching for downstream pose estimation is only responsible for finding valid matches. The specialized architectures tailored for targeted tasks with task-specific datasets significantly hinder the generalization capacity. Fortunately, pioneering studies such as \cite{pdcnet, PDCNET+, edstedt2023dkm} have endeavored to integrate sparse and dense matching within a unified architecture by jointly learning all-paired dense matching following an uncertainty-based sparsification module, leading to an appealing possibility of improved generalization capacity by taking advantage of the abundant diversity in various task-specific datasets with a unified matching model. However, the domain gap within data of different sources as well as the potential conflict in dense matching and its sparsification makes it challenging to realize this expectation.

\begin{figure}[!t]
\begin{minipage}[h]{0.57\linewidth}
    \centering	
    \includegraphics[height=3.8cm]{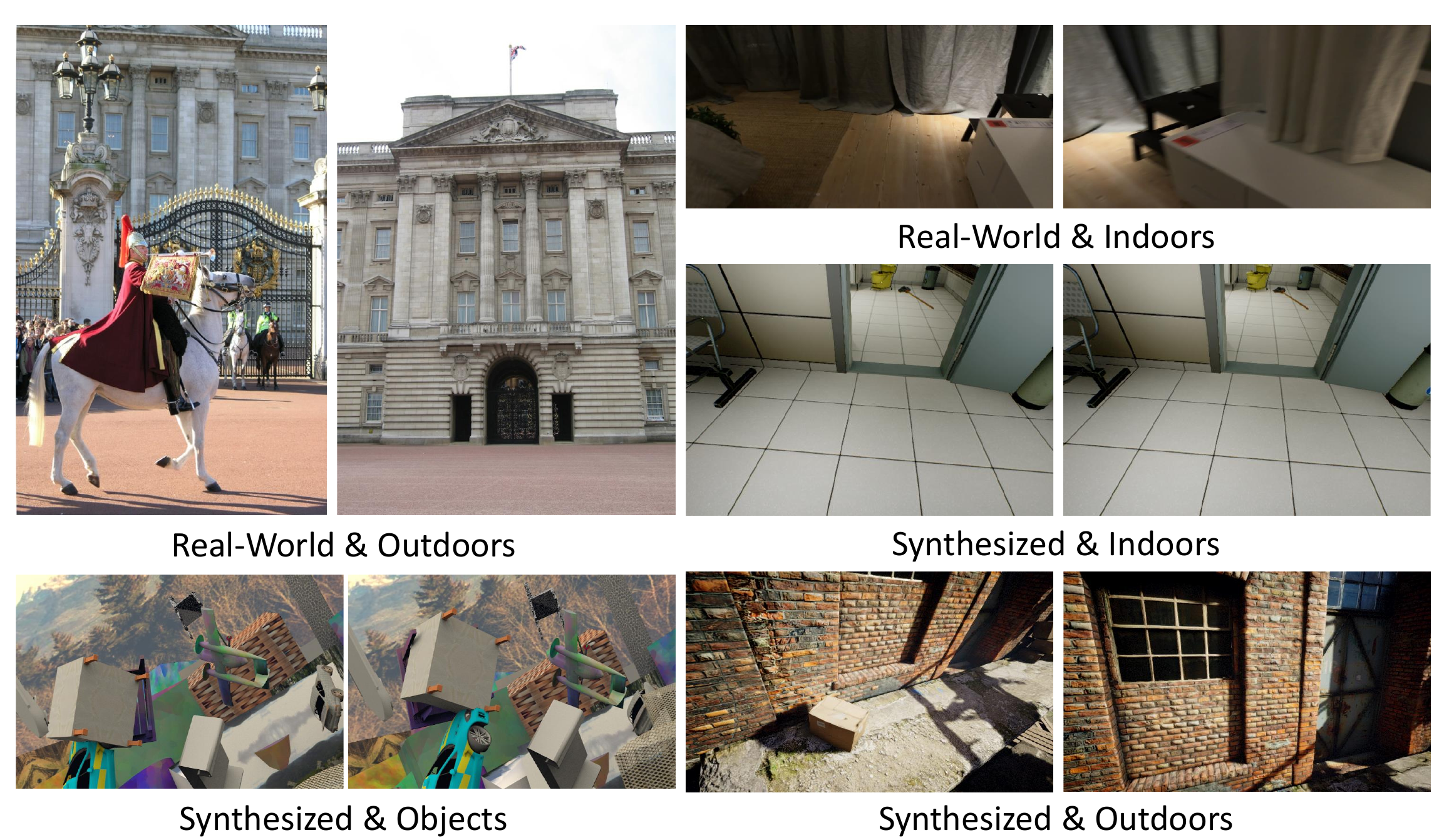}
    \tiny{(a) Overview of our training data.}
\label{fig:diverse_data}
\end{minipage}
\hfill
\begin{minipage}[h]{0.42\linewidth}
    \centering	
    \includegraphics[height=3.8cm]{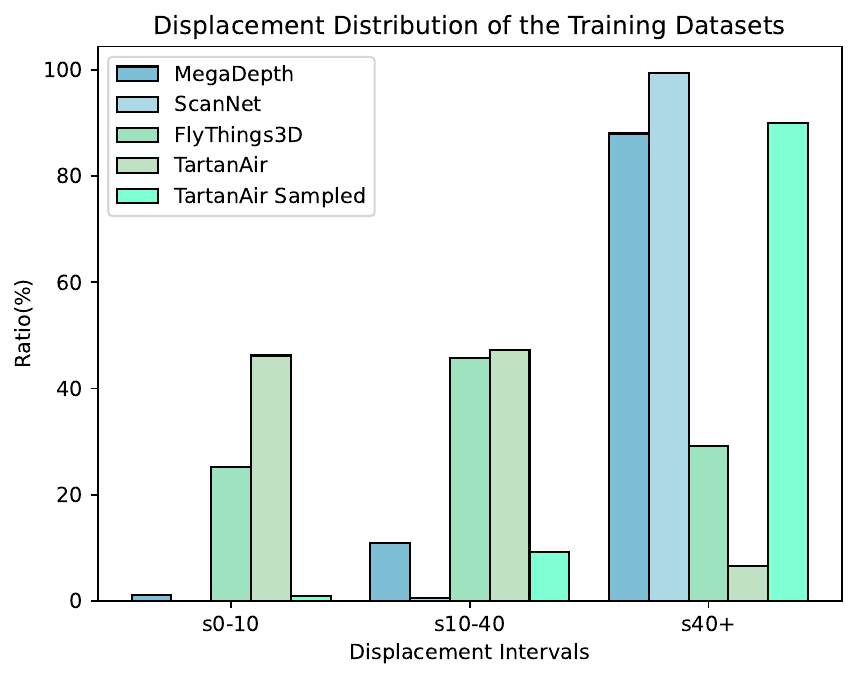}
    \tiny{(b) Displace distribution.}
\end{minipage}
\caption{\textbf{Diversity of our training data.} The collected data incorporates synthesized and real-world data covering indoor and outdoor scenarios (a) with various displacement distributions (b).}
\label{fig:data_diversity}
\vspace{-0.5cm}
\end{figure}

To tackle this challenge, we propose a \textbf{r}obust \textbf{g}eneralizable \textbf{m}atching model, termed RGM. Specifically, we decouple the learning of all-paired correspondence and uncertainty-based sparsification in a two-stage hierarchical manner by learning the uncertainty from the well-learned matches, which brings two primary advantages. The uncertainty estimation that determines the valid matches is directly based on the matching accuracy and learning from a well-learned matching model can alleviate the ambiguities caused by the erroneous correspondences, mitigating the mutual interference for better learning efficiency. Moreover, the scaling law of different datasets for different tasks can be explored separately to analyze the data conflict for better combination. To enrich the diversity of training data, datasets from tasks of dense optical flow and sparse feature matching are collected. We generate a new dataset based on the TartanAir\cite{TartanAir} with greater sampling intervals to obtain a similar displacement distribution to the real-world datasets. As shown in Fig.\ref{fig:data_diversity}, our final training data incorporate real-world and synthesized domains covering indoor and outdoor scenarios with various displacement distributions, significantly scaling up the data diversity.

Armed with our decoupled learning strategy and large-scale datasets of great diversity, our proposed RGM demonstrates excellent generalization capacity for zero-shot evaluation on matching and pose estimation. Our contributions can be summarized as follows:

Our main contributions can be summarized as follows:
\begin{itemize}
\itemsep 0cm
    \item We propose a robust, generalizable matching 
    (\OURS) network, unifying sparse and dense matching by learning in a decoupled two-stage manner which alleviates data domain gaps and potential conflict within different tasks for better learning efficiency.
    
    \item We explore collecting datasets from different tasks including dense optical flow, sparse feature matching, and a synthesized dataset whose displacement distribution is similar to the real world, significantly scaling up the diversity of training data.
    
    \item Our \OURS, for the first time, extends the training data to tasks of dense optical flow and sparse feature and achieves state-of-the-art generalization performance in \textit{zero-shot evaluation} for matching and pose estimation across multiple datasets, outperforming previous models by a considerable margin.
    
\end{itemize}

\section{Related Work}
\subsection{Sparse Feature Matching}
For a long time, solving the geometry estimation problem has been dominated by sparse correspondence matching methods. Classic methods  \cite{rublee2011orb, liu2010sift} propose robust hand-crafted local features for matching and have been adopted in many 3D reconstruction related tasks. Following the manner of detection and matching,  leaning-based methods efficiently improve the matching accuracy, among which SuperGlue
\cite{super-glue} is a representative network. Given two sets of interest points as well as the corresponding descriptors, SuperGlue utilizes a Transformer-based graph neural network for feature enhancement and to obtain great improvement. LightGlue  \cite{lindenberger2023lightglue} further modifies the SuperGlue by proposing an adaptive strategy due to the matching difficulty which effectively accelerates the inference.

Since the
introduction 
of LoFTR  \cite{loftr},  detector-free local feature matching methods, which discard the feature detector stage, 
have attracted 
great research attention  \cite{matchformer, aspanformer, tang2022quadtree}. LoFTR  \cite{loftr} takes a coarse-to-fine strategy by first establishing a dense matching correspondence and removing the unreliable matches at the refinement stage. Self-attention and cross-attention with the transformer are introduced to enlarge the receptive field. Subsequent works %
propose to 
improve upon LoFTR. %
For example, 
ASpanFormer  \cite{aspanformer} 
adopts a novel self-adaptive attention mechanism guided by the estimated flow while QuadTree   \cite{tang2022quadtree} primarily focuses on the optimization of the attention mechanism by selecting the sparse patches with the highest top $K$ attention scores for attention computation at the next level such that the computation cost can be efficiently reduced. Sparse correspondence matching plays an important role in geometry estimation 
including 
pose estimation, and 3D reconstruction.
Unfortunately, 
the sparsification of matching estimation impedes its applications when all-paired matches are required. Recently, some sparse matching works \cite{edstedt2023dkm, drcnet, pdcnet, PDCNET+} are %
proposed 
on the base of dense matching where the all-paired matching results are preserved along with a selecting module for sparsification. This is a %
milestone 
step towards unifying sparse and dense matching.

\subsection{Dense Matching}
For dense correspondence matching, 
one can 
categorize them into multiple specific tasks containing stereo matching \cite{EDNet, psmnet, AANet, crestereo, ACVNet, RAFT-Stereo}, multi-view stereo matching  \cite{RAFTMVS}, and optical flow estimation \cite{FlowNet, FlowNet2, pwcnet, raft, gmflow, flowformer, shi2023flowformer++, sui2022craft}%
. 

Dense correspondence matching is required to provide %
matching prediction per pixel even in occluded regions which are usually %
neglected 
in the sparse matching problem. Among %
dense matching, %
optical flow estimation is relatively more %
challenging 
due to the disordered motions. The 
recent 
work RAFT  \cite{raft} proposes a GRU-based iterative mechanism for refining the estimated optical flow by looking up the correlation pyramid repeatedly. %
RAFT has 
inspired a few methods for 
various dense matching tasks besides the optical flow estimation  \cite{flowformer, shi2023flowformer++, GMA, dong2023matchflow} including MVS  \cite{RAFTMVS}, and stereo matching  \cite{crestereo, RAFT-Stereo}, which validate its capacity as a universal architecture for dense matching. The limitation for dense matching lies in that only limited real-world datasets with constrained variation in perspectives are available which may hamper the generalization performance.

\subsection{Generalizable Matching}
The generalization performance has been greatly improved in monocular depth estimation \cite{midas, bhat2023zoedepth, yin2023metric} while matching models still demand further exploration. MatchFlow  \cite{dong2023matchflow} manages to improve the robustness of optical flow estimation by utilizing a model pretrained on a real-world dataset \cite{megadepth}. GMFlow~\cite{gmflow} proposes a simple yet efficient framework for computing global similarity and achieves remarkable performance on optical flow estimation. UniMatch \cite{unimatch} modifies GMFlow with additional iterative refinement and unifies the dense matching estimation for optical flow, stereo matching, and depth estimation by global matching. DKM \cite{edstedt2023dkm} and PATS \cite{pats} tackle the sparse matching tasks in the manner of dense matching, enabling them competent for unified matching. 
However, all of the above-mentioned methods are only trained on specific datasets and thus the generalization capacity is constrained when transferring to new scenarios. PDCNet and PDCNet+ \cite{pdcnet, PDCNET+} propose a universal matching framework with training on both a sparse matching dataset and a %
customized \textit{synthesized} optical flow dataset. GIM\cite{GIM} is proposed to leverage the large-scale unlabeled video sequences via a self-learning framework to improve the generalization performance.

In our work, we propose a decoupled learning strategy and gradually scale up the diversity of training data by mixing up multiple task-specific datasets from dense optical flow and sparse feature matching, significantly improving the generalization performance to a new height.

\begin{figure*}[t]
\centering
\includegraphics[width=0.9\linewidth]{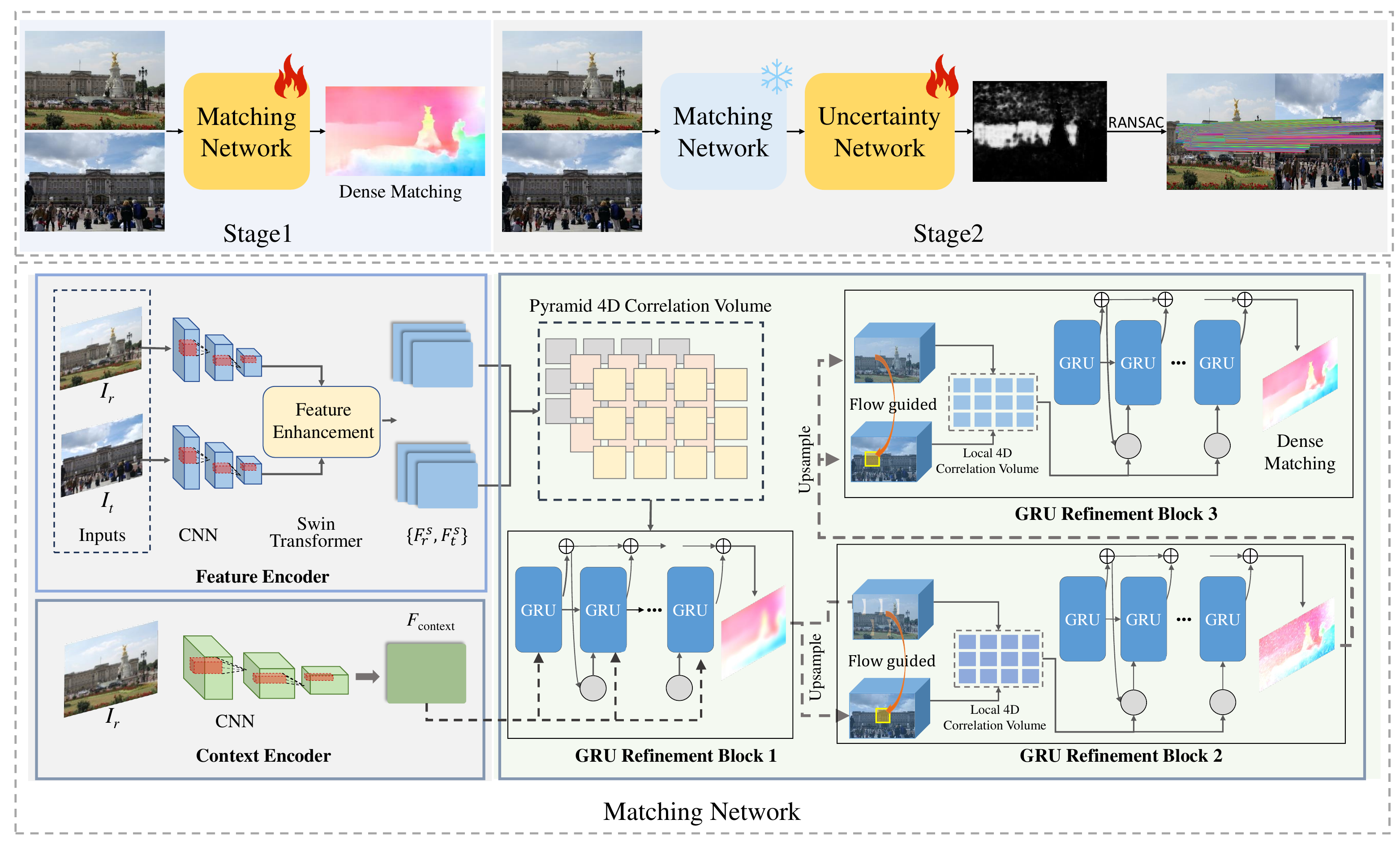}
\caption{{\bf  An overview of our proposed  \OURS}. The learning of dense matching and uncertainty-based sparsification is decoupled into two stages as shown in the upper part while the lower part illustrates the framework of our cascaded matching network.}
\label{fig:network}
\vspace{-0.1cm}
\end{figure*}

\section{Method}\label{sec:method}
In this section, we first describe the network for all-paired dense matching, then we discuss our proposed decoupled learning strategy and a synthesized dataset whose displacement distribution to the real-world images.

\subsection{Matching Network}\label{subsec:network}
Given a pair of reference and target images $\{\bI_r, \bI_t\} \in \matR^{H \times W \times 3}$, the dense matching $\bM\in \matR^{H\times W\times 2}$ is defined in the definition of optical flow. 

We follow the milestone work of RAFT \cite{raft} to find the corresponding pixels via iterative refinement. The vanilla RAFT and subsequent variants \cite{flowformer, shi2023flowformer++}, however, suffer from the utilization of a coarse feature at $1/8$ resolution which leads to an inevitable loss of fine-grained features. To alleviate this issue, we first construct a feature pyramid network to obtain a set of reference and target features $\{\mathbf{F}_r^s, \mathbf{F}_t^s\}$ at the corresponding scales $s$ of $\{1/8, 1/4, 1/2\}$ as illustrated in the lower block of Fig.\ref{fig:network}. We adopt self-attention and cross-attention based on the Swin-Transformer\cite{swin} for feature enhancement as in GMFlow \cite{gmflow} at the first two scales. 

To adapt to higher resolutions, a cascaded GRU refinement module is proposed. Instead of building all-paired correlations across every scale  \cite{crestereo, msraft}, the all-paired pyramid correlation volume $\bC_{full}$ is only introduced at the coarsest resolution of $1/8$ while the subsequent correlation volume is formulated locally. The all-paired correlation volume $\bC_{full}$ is formulated as:
\begin{align}
    \bC_{full} = \frac{\bF_r 
    \bF_t^T}{\sqrt{D}} \in \matR^{\frac{H}{8}\times \frac{W}{8} \times \frac{H}{8} \times \frac{W}{8}},
    \label{eq:global_correlation}
\end{align}
where $(H, W)$ is the spatial resolution of the original image and $D$ denotes the feature dimension.
See  GMFlow  \cite{gmflow}. A correlation pyramid is then constructed with an additional average pooling operation as RAFT. At upper scales of $\{1/4, 1/2\}$, a local correlation is computed.  Given the estimated dense matching ${\hat{\bM}}(\bf x)$ at grid $\bf x$, the local correlation with the radius $r$ is built as:
\begin{align}
    \label{eq:local_correlation}
    \bC_{local}({\bf x}) = \frac{\bF _r({\bf x})
    \bF^T_t({\mathscr N}({\bf x}+{\bV}(\bf x))}{\sqrt{D}},
\end{align}
where $\mathscr{N}({\bf x})_r$ is the set of local grids centered at $\bf x$ within radius $r$ defined as:
\begin{align}
    \label{eq:local_grid}
    \mathscr{N}({\bf x})_{r} = \{{\bf x}+{\bf dx} | {\bf dx} \in \mathbb{Z}^2, ||{\bf dx}||_1 \leq r\}.
\end{align}

Rather than initializing the hidden status at each scale \cite{msraft, crestereo}, we upsample hidden features with bilinear interpolation and pass it to the next refinement stage. Given the correlation as well as the contextual information, we compute motion features as in RAFT and feed it to cascaded GRU refinement for residual refinement which is then used for updating the matching flow iteratively. We %
use the 
$L_1$ loss for supervision across multiple scales between the matching prediction and ground truth:
\begin{align}
    L_m = \sum_{s=1}^{S}\gamma _{s} 
    \begin{cases}
        ||\hat{\bM}_s - \bM_{gt}||_1 & \text{if dense} \\
        ||\hat{\bM}_s - \bM_{gt}||_1 \odot {\bf P} & \text{if sparse}
    \end{cases},
\end{align}
where ${\bf P}$ indicates the valid mask where sparse correspondence ground truth is available and $\lambda _s$ is a scalar for adjusting the loss weight at scale $s$. We train the matching network on the mixture of optical flow datasets with per-pixel annotation and local feature matching datasets with sparse annotations.

\subsection{Decoupled Learning for Unified Matching}\label{subsec:decoupled}
Different from previous unified matching models\cite{pdcnet, PDCNET+, edstedt2023dkm} that jointly learn the dense matching and uncertainty-based sparsification simultaneously, we propose a decoupled strategy that decomposes the learning in a two-stage hierarchical manner as illustrated in the upper part of Fig.\ref{fig:network}. We argue that the joint learning strategy may introduce inevitable noise as the uncertainty estimation that determines the valid mask is closely based on the matching prediction. The matching accuracy could be inaccurate and ambiguous before the training is convergent, especially at the early stage. It is an ill-posed problem to determine valid areas given predicted matches of low quality. Besides, the scaled-up datasets from different sources exhibit significant domain gaps which require independent exploration for reasonable usage on matching and sparsification tasks. To this end, we learn to find the dense correspondence exclusively at the first stage. After the training is convergent, we freeze the matching network and start to learn the uncertainty based on the well-learned matches. 

To provide the uncertainty learning with explicit guidance, We compute the difference by warping the feature map and the RGB image of the target view to the reference view according to the estimated dense matches. Given the estimated dense  ${\hat{\bM}}(\bf x)$ from the final output of optical flow refinement at grid $\bf x$, the warped RGB difference ${\bf E}_{RGB}(\bf x)$ is :
\begin{align}
    \label{eq:rgb_diff}
    {\bf E}_{RGB}({\bf x}) = |{\bf I}_r({\bf x}) - {\bf I}_t({\bf x} + \hat{\bM}(x))| \in \matR^{H\times W \times 3}.
\end{align} 
We upsample the corresponding features at $1/2$ scale with linear interpolation and compute the warping difference as Eq.\ref{eq:rgb_diff}. Then we concatenate the differences of warped image and features and feed them to a shallow convolution network for predicting the uncertainty $\hat{\bf P} \in \matR^{H\times W}$ which is supervised the valid mask ground truth $\bf P$ with binary cross-entropy:

\begin{align}
    L_u = \sum_{grid} 
        {\bf P} \log(\hat{\bf P}) + (1-{\bf P})\log(1-\hat{\bf {P}}).
\end{align}
The uncertainty module is only applied at the sparsification stage for the downstream geometry estimation tasks. We follow the balanced sampling strategy proposed in DKM  \cite{edstedt2023dkm} to sample key points within the uncertainty threshold for pose estimation.

\subsection{Synthesized Optical Flow with Large Intervals}

\begin{figure*}[!t]
\centering	\includegraphics[width=1.01297\linewidth, 
 ]{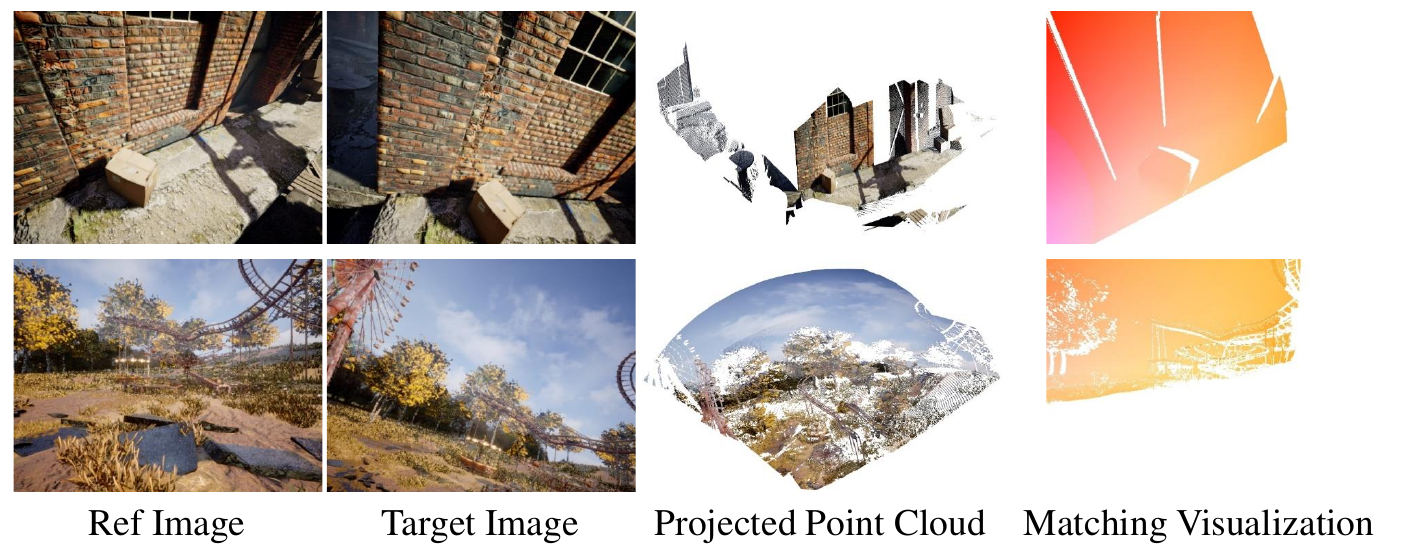}
	\caption{\textbf{Visualized comparisons.} Our  \OURS  shows superior performance by obtaining more matches given a pair of indoor or outdoor images. Moreover, our method shows the potential for robust semantic correspondence as well. The same color indicates the matched features.}
\label{fig:generated_flow}
\end{figure*}

To further exploit the advantages of fine-annotated synthetic datasets but with \textit{significant displacement}, we randomly sample frames with great intervals from 15 to 30 on the TartanAir dataset~\cite{TartanAir}. Given the provided intrinsic and extrinsic (camera-to-world) parameters as well as the depth of reference view $\{\bK_r, \bE_r, \bD_r\}$ and target view $\{\bK_t, \bE_t, \bD_t\}$. The 2D correspondences $\hat{{\bf G}}_r\in \matR^{H\times W\times 2}$ from the reference view to the target view is computed as:
\begin{align}
\mathcal{H}(\hat{{\bf G}}_r) = \bK_t \bE^{-1}_t \bE_r \bD_r \bK^{-1}_r \mathcal{H}({\bf G}_r) \in \mathbb{R}^{H \times W \times 3},
\end{align}
where the $\mathcal{H}({\bf G})\in \matR^{H\times W \times 3}$ is the homogeneous coordinates of the 2D target view's grid coordinates ${\bf G} \in \matR^{H\times W\times 2}$. The matching ground truth of the reference view $\bM_r$ is calculated as:
\begin{align}
    \bM_r = \hat{\bG}_r - \bG_r.
\end{align}
Similarly, we compute the matching ground truth for the target view $\bM_t$. Then we adopt the forward-backward consistency check~\cite{unflow, gmflow} to obtain the valid mask $\bf P$:
\begin{align}
    {\bf P} = & |\bM_r+\bM_t|^2 
    < (\alpha_1(|\bM_r|^2+|\bM_t|^2) + \alpha_2),
\end{align}
where $\alpha_1$ is $0.05$ and $\alpha_2$ is $0.5$. We synthesize around 0.7M training %
image 
pairs over $369$ scenarios to construct a new dataset named TartanAir Sampled (TS). The visualized results can be found in Fig \ref{fig:generated_flow}.

\vspace{-1mm}

\section{Experiment}
\subsection{Implementation Details}
Our proposed  \OURS  is trained in a two-stage manner. At the stage of correspondence learning, we first %
use the MegaDepth (M) \cite{megadepth} dataset with sparse correspondence ground truth. Then we 
gradually accumulate additional datasets containing the ScanNet (Sc) \cite{scannet}, FlyingThings3D (T) \cite{things3d}, TartanAir (TA) \cite{TartanAir} and our generated TartanAir Sampled(TS) datasets, reaching a total amount of 4M pairs of data. %
We follow GMFlow \cite{gmflow} with the setting of the AdamW  \cite{adamw} optimizer and the cosine learning rate schedule as well as the data augmentation for optical flow related datasets (T, TA, TS). 

At the stage of uncertainty learning, the parameters of the dense matching network are frozen. We train the uncertainty estimation module on the MegaDepth dataset for 2 epochs with a batch size of 4 and the learning rate is 1e$-4$. The training data is resized to $512\times 704$.

The related GRU iterations at $\{1/8, 1/4, 1/2\}$ scales are $\{7, 4, 2\}$ as we gradually recover the resolution for training and ablation experiments with the corresponding searching radius of $\{4, 4, 2\}$. When comparing with other approaches, the iterations increase to $\{12, 12, 2\}$ which is a closer amount of refinement compared to RAFT-based methods \cite{flowformer, raft, crestereo, sui2022craft}.

\textbf{Evaluation Metrics:} We report the average end-point-epe (AEPE, lower the better) and percentage of correct key points (PCK-T, higher the better) within a specific pixel threshold $T$. F1 metric (lower the better) is also reported for the KITTI \cite{KITTI2015} dataset which depicts the percentage of outliers averaged over all valid pixels of the dataset. For pose estimation, we follow previous works \cite{pdcnet, loftr, edstedt2023dkm, pats} by solving the essential matrix given the corresponding pixels. The accuracy is measured by AUC (higher the better) across different thresholds $(\SI{5}{\degree}, \SI{10}{\degree}, \SI{20}{\degree})$. 

\textbf{Evaluation Datasets:} To validate the generalization performance of our \OURS, we perform \textit{zero-shot evaluation} on multiple benchmark datasets containing the ETH3D \cite{eth3d}, HPatches  \cite{hpatches}, KITTI \cite{KITTI2015}, TUM \cite{tum}, and NYUD \cite{nyud} datasets for correspondence estimation. The downstream pose estimation is conducted on the TUM, NYUD, and YFCC \cite{thomee2016yfcc100m} datasets. The ground truth of matching and pose for TUM and NYUD are computed according to the provided camera parameters and depth. The geometry estimation results are also reported on the MegaDepth \cite{megadepth} datasets for ablation study. We further adopt the Sintel \cite{sintel} dataset to compare the performance of optical flow estimation.

\subsection{Zero-shot Comparisons with SOTA Methods}

\textbf{Zero-shot Correspondence Matching.} We conduct experiments to compare the generalization capacity for zero-shot matching evaluation and present the results in Table  \ref{tab:matching_comparisons}. Only sparse ground truth in valid regions is evaluated on the HPatches, TUM, ETH3D, and NYUD datasets while the occluded regions are also taken into consideration on the KITTI dataset. The specialized models \cite{raft,flowformer,unimatch} for optical flow estimation achieve relatively better performance on the KITTI dataset due to the all-paired supervision strategy and the densely annotated datasets but suffer a significant degeneration when switching to real-world scenarios as reflected by the great average end-point error and the poor percentage of correct matches. The dense matching method like DKM~\cite{edstedt2023dkm} trained on real-world data achieves a promoting accuracy on datasets except KITTI which is attributed to lacking dense supervision for occluded regions. The representative generalizable model like PDCNet+~\cite{PDCNET+} achieves an overall balance across evaluation datasets. 

Compared with the pioneering models, our RGM takes advantage of the efficient learning strategy and the abundant diversity introduced by the mixed large-scale training data from different sources, achieving the best generalization performance for zero-shot evaluation on all datasets. 

\textbf{Zero-shot Pose Estimation.}
As shown in Table  \ref{tab:pose_estimation}, our RGM also achieves the best generalization performance on YFCC (outdoors), TUM (indoors), and NYUD (indoors) datasets in terms of AUC$@\SI{5}{\degree}$ metric. This is greatly attributed to the proposed decoupled learning strategy which alleviates the interference for uncertainty-based sparsification and a reasonable combination of datasets with less domain conflict as we will discuss in Sec.  \ref{subsec:ablation_exp}.

\begin{table*}[t]
    \centering
    \caption{\textbf{Comparison with SOTA methods for zero-shot matching evaluations}. Our proposed RGM achieves the best performance among all the competing approaches. Only the \underline{underlined} results are obtained with optical flow models trained on the FlyingChairs and FlyingThings3D datasets while others are tested with additional training for Sintel submission. $^*$ indicates we utilize the officially released models and codes for evaluation at the original resolution except that we fix the maximum resolution to $920\times 1360$ for FlowFormer \cite{flowformer} due to the computation memory cost. AE is short for average end-point-error.} \label{tab:matching_comparisons}
    \setlength\tabcolsep{3pt}
    \renewcommand\arraystretch{1.3}
    \begin{tabular}{r|c|c|c|c|c|c|c|c|c|c} \hline
    \multirow{2}{*}{Method} & \multicolumn{2}{c|}{HPatches} & \multicolumn{2}{c|}{TUM} & \multicolumn{2}{c|}{ETH3D} & \multicolumn{2}{c|}{KITTI} & \multicolumn{2}{c}{NYUD} \\ \cline{2-11}
    & AE & PCK1 & AE & PCK1 & AE & PCK1 & AE & PCK1 & AE & PCK1 \\ \hline
    RAFT$^*$\cite{raft} & 60.8 & 32.5 & 8.5 & 11.6 & 6.7 & 48.7 & \underline{5.0} & \underline{17.4} & 17.5 & 47.8 \\ \hline
    FlowFormer$^*$\cite{flowformer} & 81.8 & 28.9 & 7.4 & 11.5 & 4.9 & 47.8 & \underline{4.1} & \underline{14.7} & 5.4 & 49.0 \\ \hline
    UniMatch$^*$\cite{unimatch} & 40.5 & 37.6 & 6.5 & 11.6 & 3.5 & 50.1 & \underline{5.7} & \underline{17.6} & 7.0 & 51.7 \\ \hline
    DKM*\cite{edstedt2023dkm} & 19.0 & 34.7 & 6.1 & 10.3 & 2.2 & 50.1 & 11.2 & 21.0 & 4.8 & 46.2 \\ \hline
    GLUNet$^*$\cite{GLUNet_Truong_2020} & 26.9 & 36.7 & 6.7 & 10.4 & 4.4 & 31.6 & 7.5 & 33.8 & 9.7 & 33.2 \\ \hline
    PDCNet+$^*$\cite{PDCNET+} & 17.5 & 44.9 & 4.9 & 11.5 & 2.3 & 53.3 & 4.5 & 12.6 & 4.3 & 50.0 \\ \hline
    Ours & \textbf{8.8} & \textbf{47.9} & \textbf{4.1} & \textbf{12.3} & \textbf{2.0} & \textbf{56.4} & \textbf{3.3} & \textbf{9.5} & \textbf{3.3} & \textbf{54.1} \\ \hline
    \end{tabular}
\end{table*}
\vspace{-1mm}

\begin{table}[t]
    \caption{\textbf{Downstream pose estimation}. We conduct geometry estimation on the YFCC, TUM, and NYUD datasets for zero-shot evaluations. The best results are emphasized in bold while the second best is indicated by \underline{underline}.}\label{tab:pose_estimation}
    \vspace{-1mm}
    \setlength\tabcolsep{4pt}
    \renewcommand\arraystretch{1.3}
    \centering
    \begin{tabular}{ r |c|c|c|c|c|c|c|c|c} \hline
    \multirow{2}{*}{Method} & \multicolumn{3}{c|}{YFCC (AUC)} & \multicolumn{3}{c|}{TUM (AUC)} & \multicolumn{3}{c}{NYUD (AUC)} \\ \cline{2-10}
    & @\SI{5}{\degree} & @\SI{10}{\degree} & @\SI{20}{\degree} & @\SI{5}{\degree} & @\SI{10}{\degree} & @\SI{20}{\degree} & @\SI{5}{\degree} & @\SI{10}{\degree} & @\SI{20}{\degree} \\ \hline

    LoFTR\cite{loftr} & 42.4 & 62.5 & 77.3 & 12.0 & 25.9 & 43.0 & 15.9 & 31.6 & 49.8 \\ \hline
    ASpanFormer\cite{aspanformer} & 44.5 & 63.8 & 78.4 & 13.2 & 28.2 & 46.0 & 31.4 & 49.7 & 63.3 \\ \hline
    PATS\cite{pats} & \underline{47.0} & \underline{65.3} & \underline{79.2} & 14.7 & 29.4 & \underline{46.6} & 27.4 & 46.0 & 63.9 \\ \hline
    DKM\cite{edstedt2023dkm} & 42.4 & 62.5 & 78.1 & \underline{15.5} & \underline{29.9} & 46.1 & \underline{40.6} & \textbf{58.1} & \textbf{72.8}  \\ \hline
    PDCNet+\cite{PDCNET+} & 37.5 & 58.1 & 74.5 & 11.0 & 23.9 & 40.7 & 29.5 & 48.3 & 65.9 \\ \hline
    Ours & \textbf{48.1} & \textbf{66.7} & \textbf{80.4} & \textbf{16.4} & \textbf{31.5} & \textbf{48.6} & \textbf{40.8} & \underline{57.8} & \underline{72.2} \\ \hline
    \end{tabular}
\end{table}
\vspace{-1mm}

\begin{table}[ht]
    \begin{minipage}[t]{0.48\textwidth} %
    \centering
    \caption{\textbf{Generalization Comparison with Optical Flow Models} under standard settings.}
    \vspace{-1mm}
    \renewcommand\arraystretch{1.1}
    \setlength\tabcolsep{2pt}
    \begin{tabular}{c|c|c|c|c} \hline
    \multirow{2}{*}{Method} & \multicolumn{2}{c|}{Sintel-EPE} & \multicolumn{2}{c}{KITTI} \\ \cline{2-5}
        & Clean & Final & EPE & F1 \\ \hline
        RAFT\cite{raft} & 1.4 & 2.7 & 5.0 & 17.4 \\ \hline
        GMFlow\cite{gmflow} & 1.1 & 2.5 & 7.8 & 23.4 \\ \hline
        FlowFormer\cite{flowformer} & 1.0 & \textbf{2.4} & 4.1 & 14.7 \\ \hline
        Ours & \textbf{0.9} & \textbf{2.4} & \textbf{3.9} & \textbf{12.5}  \\ \hline
    \end{tabular}
    \label{tab:optical_flow}
    \end{minipage}
    \hfill
    \begin{minipage}[t]{0.48\textwidth}
        \caption{\textbf{Generalization Comparison with DKM} trained on the Megadepth dataset. Our method obtains the best generalization performance across multiple datasets.}\label{tab:dkm_comparison}
    \vspace{-1mm}
    \setlength\tabcolsep{1pt}
    \renewcommand\arraystretch{1.3}
    \centering
    \begin{tabular}{c|c|c|c|c|c} \hline
    Method & Mega & HP & NYUD & KITTI & TUM \\ \hline
    DKM$^*$ & 3.3 & 27.1 & 14.1 & 16.2 & 20.5 \\ \hline
    Ours & \textbf{2.7} & \textbf{16.7} & \textbf{4.3} & \textbf{11.2} & \textbf{4.9} \\ \hline
    \end{tabular}
    \end{minipage}
\end{table}

\subsection{Comparison with Specialized Models}\label{ref:specialized_comparison}
We also conduct experiments following standard training settings on task-specific datasets and compare the generalization performance with SOTA methods.

\textbf{Optical Flow}. The generalization performance on the Sintel\cite{sintel} and KITTI datasets are reported after training on the FlyingChairs\cite{FlowNet} and FlyingThings3D\cite{things3d} datasets. As shown in Table  \ref{tab:optical_flow}, our RGM obtains the lowest error metrics compared with previous optical flow methods.

\textbf{Local Feature Matching}. We compare with DKM\cite{edstedt2023dkm}, the representative method for local feature matching as shown in Table  \ref{tab:dkm_comparison}. After training on the Megadepth\cite{megadepth} dataset, we report the average end-point error across multiple datasets for zero-shot matching evaluation. Our RGM surpasses the DKM with an obvious margin which is greatly attributed to our cascaded GRU refinement in the matching network.

\begin{table*}[t]
    \caption{\textbf{Ablation experiments on the decoupled training strategy}. It is clear that the independent learning strategy improves generalization performance in matching and pose estimation. M is short for Megadepth and TS is short for our generated TartanAir Sampled dataset.}\label{tab:method_ablation}
    \vspace{1mm}
    \setlength\tabcolsep{2pt}
    \renewcommand\arraystretch{1.3}
    \centering
    \begin{tabular}{c|c|c|c|c|c|c|c} \hline
    \multirow{3}{*}{Dataset} & \multirow{3}{*}{Method} & \multicolumn{3}{c|}{Correspondence Matching} & \multicolumn{3}{c}{Pose Estimation} \\ \cline{3-8}
    & & Megadepth & HPatches & ETH3D & Megadepth & NYUD & TUM   \\ \cline{3-8}
    & & AEPE & AEPE & AEPE & AUC@\SI{5}{\degree} & AUC@\SI{5}{\degree} & AUC@\SI{5}{\degree}   \\ \hline
     \multirow{2}{*}{M} & Joint & \textbf{3.5} & 31.1 & \textbf{2.3} & 53.1 & 42.4 & 14.2 \\ \cline{2-8}
     & Decoupled & 3.7 & \textbf{24.9} & \textbf{2.3} & \textbf{53.9} & \textbf{43.0} & \textbf{14.5}  \\ \hline
     \multirow{2}{*}{M+TS} & Joint & 4.0 & 18.1 & 2.1 & \textbf{52.4} & 42.3 & 14.0 \\ \cline{2-8}
     & Decoupled & \textbf{3.7} & \textbf{16.6} & \textbf{2.0} & 51.4 & \textbf{42.5} & \textbf{14.2} \\ \hline
    \end{tabular}
\end{table*}
\vspace{-1mm}

\begin{table}[t]
    \centering
    \caption{\textbf{Ablation study on the diversity for dense matching.} The zero-shot generalization capacity gradually increases as we scale up the diversity of training data.}\label{tab:data_ablation}
    \vspace{0mm}
    \setlength\tabcolsep{2.9pt}
    \renewcommand\arraystretch{1.3}
    \begin{tabular}{c|c|c|c|c|c|c|c|c} \hline
    \multirow{2}{*}{Dataset} & \multicolumn{2}{c|} {HPatches} & \multicolumn{2}{c|}{NYUD} & \multicolumn{2}{c|}{KITTI} & \multicolumn{2}{c}{TUM} \\ \cline{2-9}
    & AEPE & PCK-1 & AEPE & PCK-1 & AEPE & PCK-1 & AEPE & PCK-1 \\ \hline
    C+T & 55.3 & 38.3 & 7.6 & 52.1 & 5.0 & 72.9 & 6.1 & 12.0\\ \hline
     M & 24.9 & 44.8 & 5.3 & 48.3 & 10.8 & 71.6 & 5.2 & 11.4\\ \hline
    M+Sc & 15.3 & 42.8 & 4.0 & 50.6 & 10.4 & 71.5 & 4.2 & 12.2 \\ \hline
    M+Sc+T+TA & \textbf{13.1} & 44.3 & 3.4 & \textbf{52.7} & 3.8 & 75.0 & \textbf{4.1} & \textbf{12.3} \\ \hline
    M+Sc+T+TA+TS & 13.3 & \textbf{46.3} & \textbf{3.3} & 52.4 & \textbf{3.3} & \textbf{75.2} & \textbf{4.1} & \textbf{12.3} \\ \hline
    \end{tabular}
\end{table}
\vspace{-1mm}

\subsection{Ablation Study}\label{subsec:ablation_exp}
\textbf{Decoupled Learning Strategy:} As discussed in Sec.  \ref{subsec:decoupled}, we decouple the learning of dense matching and uncertainty-based sparsification in a two-stage manner. We conduct ablation studies on the Megadepth\cite{megadepth} and our generated TartanAir Sampled datasets to validate the effectiveness when datasets of great domain gaps and both tasks are involved. The sampling strategy in DKM\cite{edstedt2023dkm} for key points selection is not adopted at this stage to reduce affecting factors. As shown in Table  \ref{tab:method_ablation}, when learning the dense matching and uncertainty for sparsification in a decoupled manner on the Megadepth dataset, the matching generalization performance benefits an obvious improvement on the HPatches\cite{hpatches} dataset, decreasing the AEPE from 31.1 to 24.9. Moreover, the AUC metric for pose estimation also improves on the NYUD\cite{nyud} and TUM\cite{tum} datasets, from 42.4 to 43.0 and 14.2 to 14.5, respectively. When the synthesized data of our TartanAir Sampled dataset is further introduced, although the matching generalization performance obtains overall promotion on HPatches and ETH3D\cite{eth3d} datasets, it is clear that our decoupled mechanism provides better learning efficiency. Although the pose estimation on the Megadepth suffers a drop, better generalization performance can be obtained on the NYUD and TUM datasets with our decoupled strategy. 

It's worth noticing that the geometry estimation suffers an overall degeneration when introducing the synthesized TartanAir Sampled dataset, indicating the potential data conflict for sparsification which will be discussed in the following.

\textbf{Scaling Up Diversity for Matching}. We adopt the decoupled training strategy discussed above and train the dense matching network with mixed datasets. It is clear from Table  \ref{tab:data_ablation} that as we scale up the diversity of training data, the overall generalization performance of correspondence matching improves gradually. Due to the synthesized training data  \cite{FlowNet, things3d} and limited changes of viewpoint, the poor generalization performance from the optical flow estimation is %
expected which is reflected by the high AEPE across multiple datasets.

As the evaluation datasets are all real-world scenarios, the accuracy is clearly improved as the training datasets switch to the real-world MegaDepth except on the KITTI dataset which requires dense matching estimation including the occluded regions. We then fine-tune the model on the mixture of different datasets for 1 epoch separately. The introduction of the real-world indoor ScanNet 
dataset brings overall improvements on the HPatches  \cite{hpatches} and ETH3D \cite{eth3d} datasets indicated by the decreasing average end-point error.

When the optical flow datasets are further mixed at the fourth line, the average end-point error and percentage of outliers accordingly drop from 10.8 to 4.1 and $17.7\%$ to $10.8\%$ on the KITTI dataset  \cite{KITTI2015}, and the PCK-1 metric goes up to $55.9\%$ on the ETH3D datasets. The performance on the HPatches dataset obtains further boost as the AEPE metric reaches the lowest result. This improvement significantly emphasizes the importance of collecting fine-annotated optical flow datasets to improve diversity.

After the employment of our synthesized dataset, the PCK-1 metric reaches $46.3\%$ on the HPatches and the percentage of outliers decreases to $9.6\%$ on the KITTI dataset, surpassing the baseline by 47.0\%. 

\textbf{Scaling Up Diversity for Sparsification}. As mentioned in the ablation study for decoupled learning strategy, data from different sources may impede the generalization performance for learning the uncertainty. We conduct experiments on the Megadepth, ScanNet, and our generated TartanAir Sampled datasets to explore this data conflict. As shown in Table  \ref{tab:uncertainty_ablation}, the best overall generalization performance comes from the Megadepth dataset. The generalization performance on the TUM and NYUD datasets remains relatively consistent but suffers an obvious degeneration on the YFCC dataset. The introduction of the synthesized dataset, however, directly leads to a drop over all three datasets in terms of AUC$@\SI{5}{\degree}$ metric. This reveals the interference caused by significant domain gaps. Synthesized data typically shares common characteristics, including camera parameters, brightness, contrast, \etc, while images from the real world may display significant variations in these aspects. The experiments in Table  \ref{tab:uncertainty_ablation} suggest that the diversity in training data may introduce unnecessary interference to the uncertainty-based sparsification. Consequently, our final sparsification network is trained on the Megadepth dataset for overall generalization performance.

\begin{table}[t]
    \caption{\textbf{Ablation study on the diversity for uncertainty-based sparsification.} The introduction of synthesized data leads to an obvious degeneration for pose estimation, indicating the data conflict for this task.}\label{tab:uncertainty_ablation}
    \vspace{-1mm}
    \setlength\tabcolsep{4pt}
    \renewcommand\arraystretch{1.3}
    \centering
    \begin{tabular}{c|c|c|c|c|c|c|c|c|c} \hline
    \multirow{2}{*}{Dataset} & \multicolumn{3}{c|}{YFCC (AUC)} & \multicolumn{3}{c|}{TUM (AUC)} & \multicolumn{3}{c}{NYUD (AUC)} \\ \cline{2-10}
    & @\SI{5}{\degree} & @\SI{10}{\degree} & @\SI{20}{\degree} & @\SI{5}{\degree} & @\SI{10}{\degree} & @\SI{20}{\degree} & @\SI{5}{\degree} & @\SI{10}{\degree} & @\SI{20}{\degree} \\ \hline
    M & \textbf{48.1} & \textbf{66.7} & \textbf{80.4} & 16.4 & 31.5 & \textbf{48.6} & \textbf{40.8} & \textbf{57.8} & \textbf{72.2} \\ \hline
    M+Sc & 47.6 & 66.1 & 80.0 & \textbf{16.6} & \textbf{31.6} & 48.4 & 40.6 & 57.7 & 72.1 \\ \hline
    M+Sc+TS & 46.2 & 64.9 & 78.9 & 16.4 & \textbf{31.6} & \textbf{48.6} & 39.8 & 57.1 & 72.0 \\ \hline
    \end{tabular}
\end{table}
\vspace{-1mm}

\begin{figure*}[!t]
\centering	\includegraphics[width=0.98\linewidth]{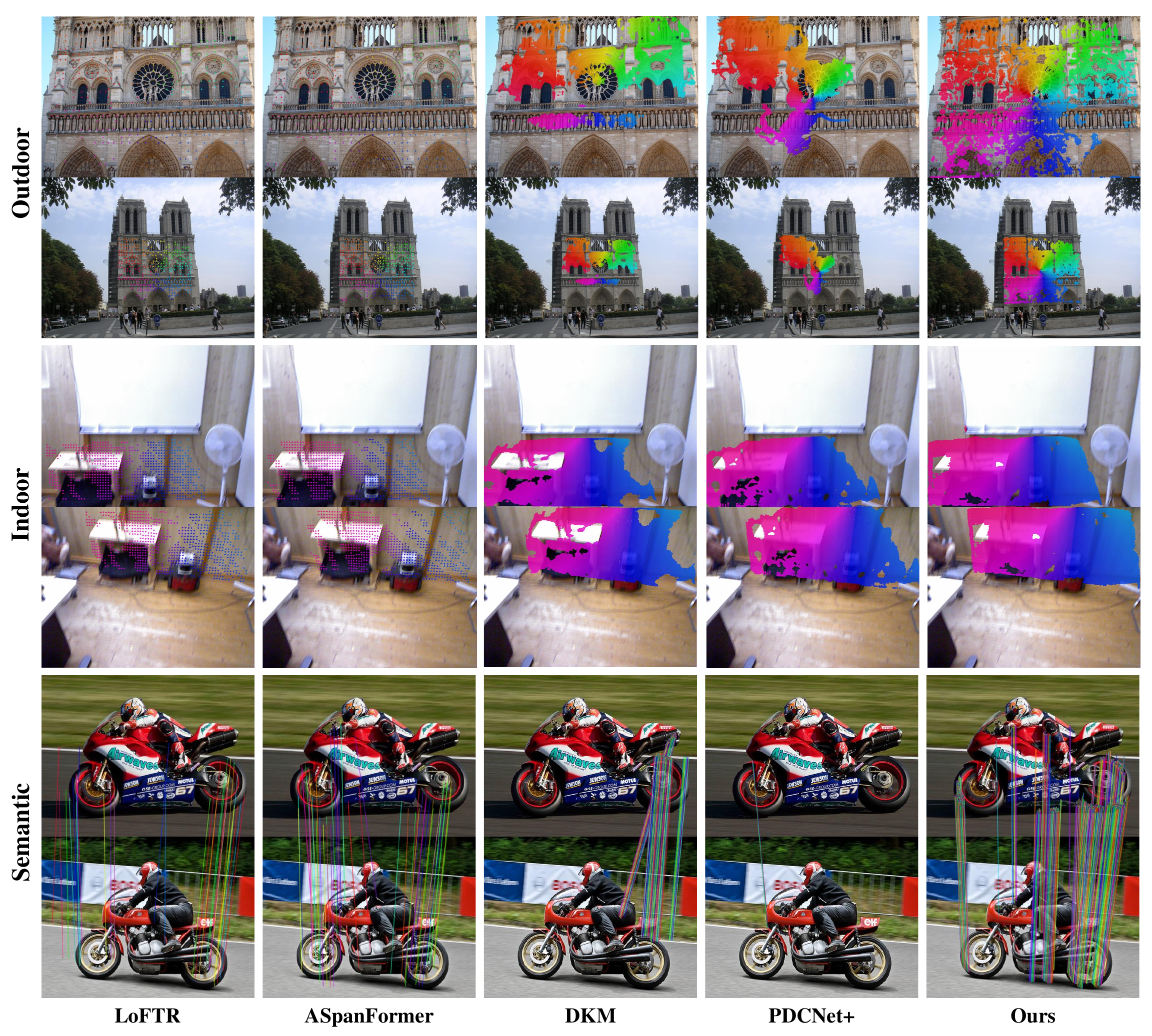}
\caption{\textbf{Visualized comparison.} Given a pair of indoor or outdoor images, our \OURS exhibits outstanding generalization performance as denser matches are obtained where the same color denotes the identical correspondence. Moreover, our method shows the potential for robust semantic correspondence as well.}
\label{fig:vis_comparisons}
\end{figure*}
\vspace{-1mm}

\section{Limitation}
Given the uncertainty estimation for sparsification, it is equally important to select the key points for pose estimation. In this paper, we follow the DKM\cite{edstedt2023dkm} with their balanced sampling strategy. We will explore this part in the future.

\section{Conclusion}
In this paper, we propose a robust and generalizable matching model capable of sparse and dense matching termed \OURS. A decoupled learning strategy is explored to decompose the dense matching and its uncertainty-based sparsification into a two-stage manner, which helps alleviate the interference introduced by task conflict and significant domain gaps within various datasets. The diversity of training data is greatly scaled up by collecting datasets from tasks of dense optical flow, sparse local feature matching, and a synthesized dataset with a similar displacement distribution to real-world images. Our RGM demonstrates state-of-the-art generalization performance on zero-shot matching and geometry estimation, outperforming previous methods by a large margin.

\bibliographystyle{splncs04}
\bibliography{main}
\end{document}